# Semantic Stereo for Incidental Satellite Images

Marc Bosch, Kevin Foster, Gordon Christie, Sean Wang, Gregory D Hager, and Myron Brown

The Johns Hopkins University
myron.brown@jhuapl.edu

## Abstract

*The increasingly common use of incidental satellite images for stereo reconstruction versus rigidly tasked binocular or trinocular coincident collection is helping to enable timely global-scale 3D mapping; however, reliable stereo correspondence from multi-date image pairs remains very challenging due to seasonal appearance differences and scene change. Promising recent work suggests that semantic scene segmentation can provide a robust regularizing prior for resolving ambiguities in stereo correspondence and reconstruction problems. To enable research for pairwise semantic stereo and multi-view semantic 3D reconstruction with incidental satellite images, we have established a large-scale public dataset including multi-view, multi-band satellite images and ground truth geometric and semantic labels for two large cities. To demonstrate the complementary nature of the stereo and segmentation tasks, we present lightweight public baselines adapted from recent state of the art convolutional neural network models and assess their performance.*

## 1. Introduction

Multi-view stereo methods for 3D mapping of the Earth increasingly rely on large archives of incidental satellite images which are more plentiful than well controlled binocular or trinocular coincident image sets which are more time-consuming and expensive to collect. Facciolo et al. [1] have demonstrated that 3D reconstruction from well selected multi-date satellite image pairs can approach the accuracy of stereo from coincident image collection. However, stereo correspondence remains a challenging problem for incidental image pairs with significant appearance differences due to seasonal change (Figure 1). When the pool of available image pairs is shallow, more robust matching methods are required.

Two of the most promising recent research directions in computational stereo are pairwise stereo matching methods using convolutional neural networks (CNNs) [2] and the incorporation of semantic labels into stereo matching and 3D reconstruction methods [3]. Zhu et al. [4] survey recent progress and discuss implications for remote sensing. Deep learning methods incorporating semantic priors appear particularly well suited for stereo correspondence with incidental satellite image pairs which often have significant appearance differences. Several recent public challenge datasets focusing on satellite images have been instrumental in benchmarking and advancing the state of the art for semantic labeling tasks [5, 6, 7, 8] and multi-view 3D reconstruction [9, 10]. To date there is to our knowledge no public labeled dataset or evaluation methodology for promoting machine learning research to exploit semantic and stereo cues together for remote sensing.

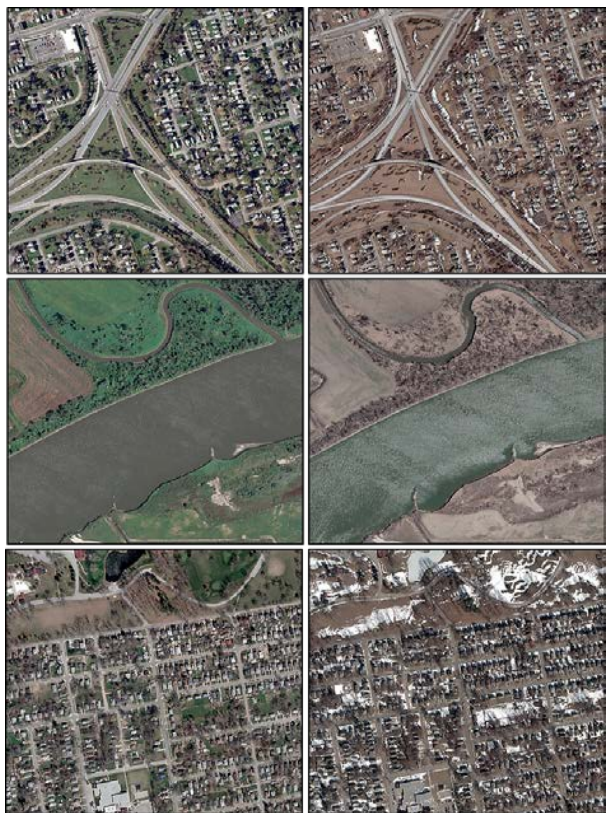

Figure 1: Seasonal appearance differences pose challenges for stereo correspondence

In this work, we present Urban Semantic 3D (US3D), a large-scale public dataset including multi-view, multi-band satellite images and ground truth geometric and semantic labels for two large cities. This dataset provides training and validation data to enable research for pairwise semantic stereo and multi-view semantic 3D reconstruction for the very challenging case of incidental satellite image collection with significant appearance differences due to seasonal change. This dataset also enables research for semantic lidar point cloud classification [11] and single image height estimation from remote sensing images [12, 13, 14]. We examine both state of the art heavyweight convolutional neural network architectures and lightweight real-time architectures for stereo correspondence and semantic image segmentation and present baseline methods and an evaluation methodology for pairwise semantic stereo and multi-view semantic 3D reconstruction that show the complementary nature of the stereo and segmentation tasks. The dataset and baseline algorithm implementations will be made publicly available.

## 2. Related Work

**Stereo Benchmarks:** Public benchmark datasets and leaderboards for stereo evaluation such as Middlebury [15], KITTI [16], and more recently ETH3D [17] have been instrumental in advancing the state of the art. In recent years, data-driven machine learning methods have increasingly dominated the leaderboards. The 2016 Multi-View Stereo 3D Mapping (MVS3DM) challenge produced a large public dataset for evaluating stereo algorithms applied to satellite images [9]. The top performing solution was based on Facciolo et al.'s More Global Matching (MGM) method [18] which refines Hirschmuller's Semi-Global Matching (SGM) method first proposed in 2005 [19]. The US3D dataset provides labeled training and validation data to also enable machine learning research for stereo with satellite images.

**Semantic Segmentation Benchmarks:** Thanks to public competitions such as the Pascal Visual Object Classes (VOC) Challenge [20], Microsoft Common Objects in Context (COCO) [21], Cityscapes [22], and Mapillary Vistas [23], semantic segmentation algorithm performance has improved dramatically in recent years. Public competitions such as SpaceNet [5], Urban 3D Challenge [6, 7], and DeepGlobe [8] have also been instrumental in benchmarking semantic segmentation algorithms for classifying buildings in satellite images. The US3D dataset extends this work by providing semantic labels for buildings, ground, trees, and water.

**Semantic Stereo Benchmarks:** Public data including stereo disparity and semantic labels for street level scenes is available in the KITTI [16], CityScapes [22], and SYNTHIA [24] datasets, enabling research focused primarily on autonomous driving [25, 26, 27, 28, 29]. The recent 3D Reconstruction Meets Semantics (3DRMS) workshop provided labeled data focused on navigation of natural environments [30]. With the US3D dataset, we seek to leverage lessons learned from these applications and apply them to remote sensing tasks.

## 3. The US3D Dataset

The US3D dataset currently includes approximately 100 square kilometer coverage for the United States cities of Jacksonville, Florida and Omaha, Nebraska, as shown in Figure 2. We leverage source data that has recently been made publicly available. Sources include incidental satellite images, airborne lidar, and feature annotations derived from lidar. Our data curation process is entirely automated except for manual editing of small subsets of data, primarily reserved for algorithm fine-tuning and sequestered testing. This will enable expansion of the US3D dataset to include other cities as additional data becomes available.

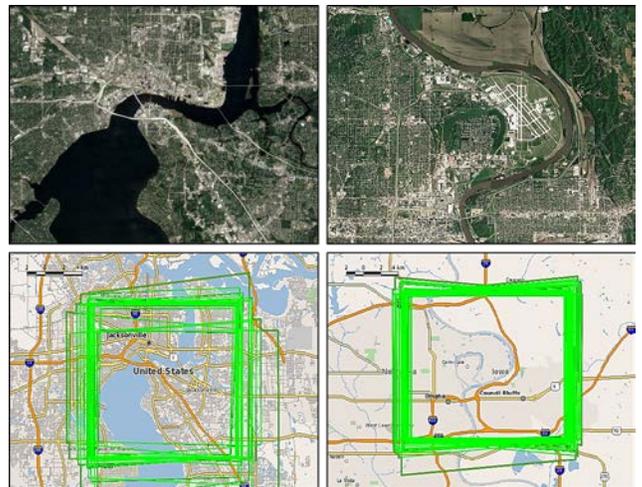

Figure 2: Areas covered by WorldView-3 satellite images of Jacksonville (left) and Omaha (right) shown in Google Earth (top) and DigitalGlobe ImageFinder (bottom)

### 3.1. Incidental Satellite Images

Source imagery for this dataset is provided courtesy of DigitalGlobe. Twenty-six WorldView-3 target-mode panchromatic, visible, and near infrared (VNIR) images collected between 2014 and 2016 over Jacksonville, Florida and forty-three WorldView-3 target-mode panchromatic and VNIR images collected between 2014 and 2015 over Omaha, Nebraska are provided in National Imagery Transmission Format (NITF) with Rational Polynomial Coefficient (RPC) sensor model metadata. The areas imaged are shown in Figure 2 and cover approximately 100 square kilometers. Panchromatic ground sample distance (GSD) is approximately 30cm and VNIR GSD is approximately 1.3m. The distribution of image collection months as shown in Figure 3 is sufficient to enable stereo performance evaluation for a broad range of seasonal differences.

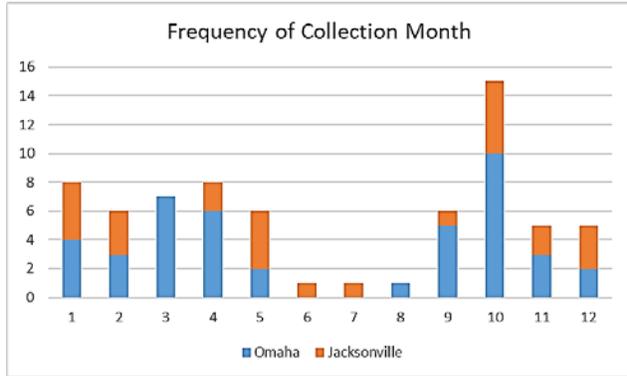

Figure 3: Distribution of image collection months

### 3.2. Airborne Lidar

Ground truth geometry for stereo evaluation is provided by airborne lidar collected over 133 U. S. cities and previously made publicly available by the Homeland Security Infrastructure Program (HSIP) [31]. The lidar aggregate nominal pulse spacing (ANPS) is 80cm, roughly equivalent to two or three pixels in the satellite imagery. This data is intended for training and validation.

In addition to HSIP lidar for the entire dataset, we are also acquiring lidar and oblique airborne imagery from Geomni for 16 square kilometers over each site. The lidar ANPS will be approximately 25cm, less than one pixel in the satellite imagery. This data is intended for fine-tuning and sequestered testing. The airborne images will have approximately 5cm GSD, suitable for then assessing high-resolution 3D reconstruction methods.

### 3.3. Semantic Labels

Semantic labels for the US3D dataset are derived automatically from publicly available HSIP lidar products (see Figure 4). In addition to lidar point clouds, the HSIP U. S. Cities dataset includes Digital Surface Model (DSM) and Digital Terrain Model (DTM) products, tree center points, and building polygons automatically produced using the lidar source data and manually edited to ensure reasonable accuracy. Building polygons from the HSIP dataset were employed as ground truth labels for building classification in the 2017 Urban 3D Challenge [7], and we here used the additional data provided to label ground, buildings, trees, and water. There is class imbalance in the tiled dataset as shown in Figure 5, and individual tiles with little semantic diversity (e.g., over water) were removed.

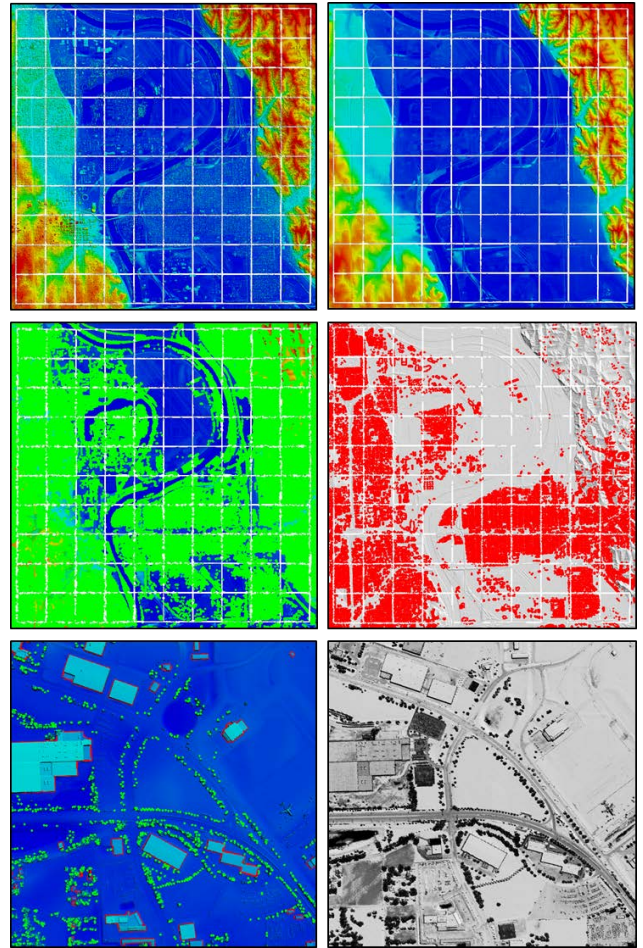

Figure 4: Tiled public lidar DSM and DTM (top), tree center points and building polygons (center), and individual DSM and lidar intensity image tiles with semantic labels (bottom)

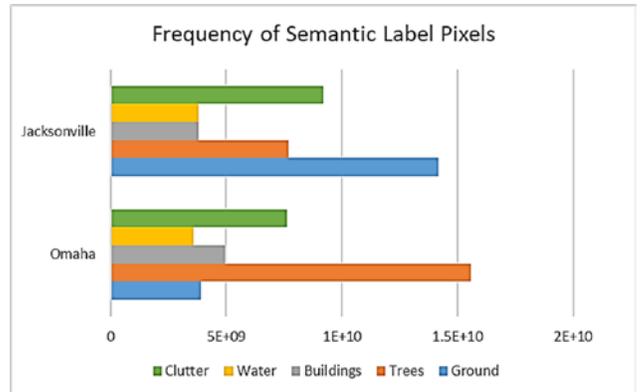

Figure 5: Distribution of semantic labels in the dataset

### 3.4. Labeled Satellite Image Tiles

Satellite images are provided in geographically non-overlapping tiles, with orthorectified semantic labels, XYZ coordinates, and normalized DSM above-ground heights all projected into each image plane. Each image tile is aligned with lidar intensity using the mutual information metric and the RPC line and sample offset metadata are updated. Solar angle metadata is exploited to cast shadows in the lidar intensity image to improve reliability of image matching. Pan-sharpened images are produced for both three-channel RGB and eight-channel VNIR. All sensor metadata is retained in each tiled image.

To simplify the pairwise stereo task, we also provide epipolar rectified image pairs and ground truth disparity and semantic labels produced using the tiled images. Since adjusted RPC sensor model coefficients are available for each image tile, we determine virtual correspondences for epipolar rectification by projecting a range of xyz world coordinates into each pair of images as described by de Franchis [32]. This is very efficient and avoids degenerate configurations of correspondences in low relief terrain. Similarly, we produce ground truth image pair disparities by projecting known xyz coordinates from each left image tile into the left and right epipolar rectified images. Any residual y parallax is then measurable and provides a check on the accuracy of the adjusted RPC sensor models. Metadata such as RPC, epipolar rectifying homographies, and collection dates are retained for each stereo pair. An example stereo pair with ground truth disparities and semantic labels is shown in Figure 6.

## 4. Benchmark Tasks and Metrics

### 4.1. Pairwise Semantic Stereo

The first benchmark task is pairwise semantic stereo. Given pairs of epipolar rectified images, the objective is to produce accurate semantic labels and stereo disparity estimates. The two subtasks may be considered separately or sequentially and then fused [26, 27], semantic segmentation may be utilized as a prior for stereo [25, 28], stereo may be utilized as a prior for semantic segmentation [33], or both may be solved together [3, 29]. Image pairs with a broad range of temporal differences are included for assessing algorithm robustness to severe seasonal appearance differences.

Stereo performance is assessed using average endpoint error (EPE) and fraction of erroneous pixels (D1) for consistency with Middlebury [15] and KITTI [16] evaluations. For D1, pixels with disparity errors greater than three pixels are considered erroneous to account for inaccuracies in the ground truth lidar data which has approximately 80cm point spacing compared to 30cm satellite images. Semantic segmentation performance is assessed using mean intersection over union (mIoU) for consistency with COCO [21]. Semantic stereo is also assessed using a combined mIoU score for which true positives have both correct semantic label and disparity error less than three pixels.

### 4.2. Multi-View Semantic 3D Reconstruction

The second benchmark task is multi-view semantic 3D reconstruction. Multiple epipolar rectified image pairs are provided for each geographic tile to enable multi-view stereo by combining pairwise solutions [1, 34]. Unrectified images with RPC metadata, semantic labels, and ground truth XYZ and normalized height coordinates for each pixel are also provided to enable more general solutions [34]. For assessment of complete stereo pipelines, the source NITF images are also provided along with orthorectified ground truth semantic labels and UTM and height coordinates. One of the significant challenges in multi-view stereo is reliable image pair selection based on known imaging geometry [35] and temporal proximity [1]. Rectified and unrectified images in the US3D dataset retain all NITF metadata required for these purposes.

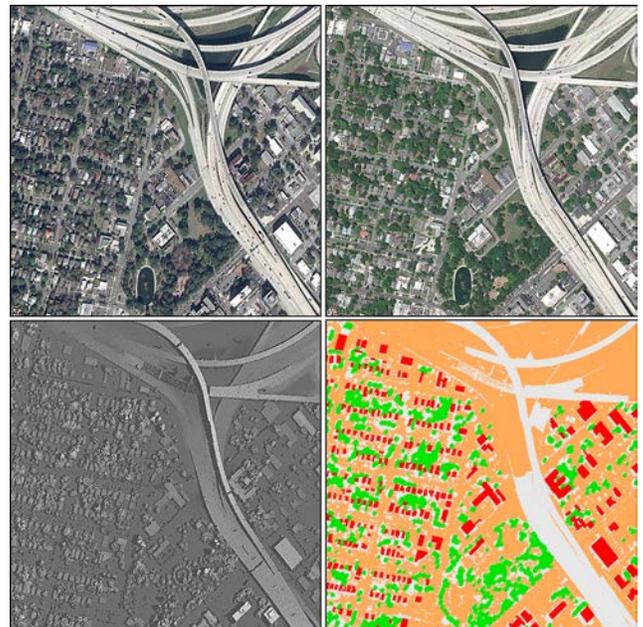

Figure 6: Example epipolar rectified images (top) with ground truth left disparities and semantic labels (bottom)

Multi-view semantic 3D reconstruction performance is assessed using Z accuracy and completeness metrics for stereo [9, 10] and a combined mIoU score for which true positives have both correct semantic label and Z error less than one meter. For assessing end-to-end pipelines for multi-view stereo using NITF source images, a 3D translation must first be estimated to account for any inaccuracies in bundle adjustment.

### 4.3. Single-View Height Estimation

There has so far been little work in estimating heights from single-view remotely sensed images. Mou and Zhu [14] recently proposed a CNN trained solely with DSM height. Srivastava et al. [13] proposed joint estimation of semantic labels and normalized DSM height. The US3D dataset provides DSM, DTM, and nDSM heights along with image collection metadata useful for more fully exploring and evaluating these and other ideas.

### 4.4. Point Cloud Semantic Segmentation

For each of the geographic tiles, lidar data employed for assessing stereo algorithms is also provided in point cloud form such that each 3D point is classified with a semantic label. This enables development and evaluation of point cloud semantic segmentation algorithms. We borrow the conventions for both file format and scoring methodology from SEMANTIC3D.NET [11]. Performance is assessed using mIoU and overall accuracy (OA).

## 5. Baseline Methods

To validate the utility of the presented data set and to demonstrate the complementary nature of the stereo and segmentation tasks, we have adapted and assessed both heavyweight state of the art convolutional neural network architectures and lightweight architectures designed for real-time applications. The real-time models are especially useful for efficient experimentation. All of the baseline implementations will be provided along with the publicly available data set.

### 5.1. Stereo Correspondence

For pairwise stereo correspondence, we first considered a Semi-Global Matching (SGM) [19] implementation that incorporates weighted least squares (WLS) edge-aware disparity map filtering [36]. This is available in OpenCV and runs quickly, providing a useful quick indicator of the challenges observed in each epipolar rectified image pair.

Data-driven CNN based methods now significantly outperform most conventional stereo algorithms such as SGM on public benchmarks, so we trained and evaluated two recently proposed models with the US3D dataset – Liang et al.'s state of the art iResNet [37] and Atienza's remarkably small real-time DenseMapNet [38].

The iterative residual prediction network (iResNet) was introduced as an architecture that includes four steps common in a stereo matching problem – cost calculation, cost aggregation, disparity estimation, and disparity refinement in an end-to-end trainable framework. The architecture consists of three parts: (1) multi-scale feature extraction used to calculate the cost, (2) an encoder-decoder subnetwork to compute initial disparity estimates, and (3) a refinement subnetwork that takes the initial disparities and two feature constancy terms and computes residual disparities that are added to the initial estimates producing the final disparity values.

DenseMapNet combines a correspondence network to learn stereo matching and a disparity network to directly regress disparity values for the reference image. The architecture is inspired by DenseNet [39] which concatenates each layer to the subsequent layer such that the loss function has direct access to all feature layers. Concatenated layers share weights, leading to a compact model requiring only 290k parameters so it is easy to train from scratch. Multiple scales of dilated convolutions [40] are used to incorporate context into the correspondence network and to enable increased maximum disparity search.

We adapted DenseMapNet in the following ways for our purposes. First, we removed disparity normalization to enable direct regression of bidirectional disparities. We then implemented a custom mean squared error loss function that ignores unlabeled ground truth disparity values since some scene feature disparities (e.g., water) are not measurable. Finally, we replaced 2D convolutions with depth-wise separable convolutions as proposed by Chollet [41] to further reduce the number of parameters to 90k without sacrificing accuracy. While the concatenated layers in DenseMapNet share weights, the current Keras implementation with TensorFlow backend does not share memory allocation as proposed by Pleiss et al. [42]. Even so, the current implementation performs well enough on a GTX 1070 with 7GB of GPU memory.

### 5.2. Semantic Segmentation

Several semantic segmentation algorithms have been proposed in recent times, a majority of which consist of a backbone neural network (e.g., MobileNet, ResNet, or VGG) to extract features combined with higher level layers that perform pixel level inference. A popular model is DeepLab [41, 43] which leverages atrous or dilated convolutions and spatial pyramid pooling layers. Atrous convolutions help simulate the effect of differently sized receptive fields without adding new parameters into the model as features are extracted. The pyramid pooling layers consist of parallel atrous convolutional layers that process feature maps, each with a different sample spacing parameter. This provides a powerful multi-resolution processing mechanism.

For a small baseline model, we trained Zhao et al.'s Image Cascade Network (ICNet) [44] which fuses feature maps from multi-resolution branches, each guided by a supervised loss. The lowest resolution branch is a fully convolutional Pyramid Scene Parsing Network (PSPNet) [45] with dilated convolutions to enlarge the receptive fields. Medium and high resolution branches each include three convolutional layers with down-sampling at each layer. The highest resolution branch is 4x down-sampled, and up-sampling is applied for inference.

We adapted ICNet in the following ways for our purposes. First, we replaced the cross entropy loss for each of the cascade branches with a custom Jaccard loss to better handle underrepresented classes, and we ignore unlabeled pixels in the loss function. We then generalized the model inputs to support arbitrary numbers of image bands, though our reported experiments do not incorporate the 8 channel multispectral images. Finally, we replaced 2D convolutions with depth-wise separable convolutions to reduce the parameter count from 6.74M to 3.82M for three channel input images without sacrificing accuracy.

### 5.3. Semantic Stereo

To demonstrate the value of semantic category as a prior for stereo correspondence, we adapted DenseMapNet to include four input channels with the fourth channel being a classification label from semantic segmentation. This simple approach facilitates sequential execution and bootstrapping of these complementary tasks.

### 5.4. Multiple View Semantic Stereo

For our baseline multiple view semantic stereo method, we adapt the straightforward median filter approach demonstrated for stereo by Ozcanli et al. [34], Facciolo et al. [1], and Bosch et al. [9]. For each geographic tile, we perform pairwise semantic stereo, triangulate to world coordinates in the ground plane, apply median filters to both reconstructed 3D heights and associated semantic labels. For efficient triangulation of correspondences, we approximate each 80 coefficient RPC sensor model with a 3x4 projection matrix for each 2048 x 2048 pixel image tile. Approximation error was confirmed to be less than one tenth of a pixel, similar to results shown by de Franchis et al. [32]. This avoids the complexity of more general inverse solvers [46, 47].

## 6. Experimental Evaluation

We have conducted initial experiments using our baseline methods to verify correctness of the dataset, to set expectations for performance, and to demonstrate the value of semantic cues for improving stereo. For these experiments, we randomly sampled 1,000 image pairs from the dataset which is less than one percent of the total pool, not including image tiles set aside for sequestered testing. Eighty percent of the image pairs are used to training and twenty percent for validation. We demonstrate results using the RGB images.

DeepLab and iResNet were trained and evaluated on a single Quadro M6000 with 24 GB of GPU memory. ICNet and DenseMapNet were trained on a GTX 1070 with 8 GB of GPU memory. We exploited gradient checkpointing [48] to reduce memory allocation for backpropagation, allowing us to double the mini-batch sizes for training on a laptop. Model sizes and run times are shown in Table 1.

Semantic segmentation baseline results are shown in Table 2. The mIoU scores for both DeepLab and ICNet are reasonable and help to confirm the correctness of the dataset; however, baseline performance leaves significant room for improvement. The semantic categories are very general with significant in-class variation, and the small subset of the data used for fine-tuning DeepLab and training ICNet is insufficient to capture the variety of appearance. The baseline methods also do not incorporate geometric priors from stereo which has been shown to improve performance.

Table 1: Model sizes and run times for 1024x1024 pixel images

| Method | Model Parameters | Time (ms) Quadro M6000 | Time (ms) GTX 1070 |
|---|---|---|---|
| SGM + WLS | - | - | 170 |
| iResNet-i2 | 43.3M | 217 | - |
| DenseMapNet | 0.1M | - | 97 |
| DeepLab v3 | 58M | 1517 | - |
| ICNet | 3.8M | - | 59 |

Table 2: Semantic segmentation performance

| Method | mIoU |
|---|---|
| ICNet | 0.70 |
| DeepLab v3 | 0.75 |

Table 3: Pairwise stereo and semantic stereo performance for all image pairs, including those with significant label noise

| Method | EPE | D1 |
|---|---|---|
| SGM + WLS | 10.34 | 0.43 |
| iResNet-i2 (pre-trained) | 6.31 | 0.38 |
| DenseMapNet | 3.49 | 0.34 |
| iResNet-i2 (fine-tuned) | 3.05 | 0.33 |
| DenseMapNet + Truth CLS | 3.51 | 0.35 |

Pairwise stereo baseline results are shown in Table 3. As expected, both iResNet and DenseMapNet significantly outperform SGM, eliminating the most egregious outliers. Interestingly, the tiny DenseMapNet nearly performs as well as iResNet with two orders of magnitude fewer parameters. The fraction of erroneous pixels D1 is very high. The stereo matching problem here is extremely challenging, with significant appearance change observed in the image pairs as shown in Figure 1. We trained and tested the simple four-channel semantic stereo method using ground truth classification (CLS) labels and initially observed no obvious benefit. Some of the appearance changes observed in incidental image pairs are due to either seasonal geometric change such as leaf-on and leaf-off foliage or man-made change such as transient vehicles and construction. To address this for a large scale dataset for training and validation without explicit manual image pair selection, we reviewed SGM endpoint error scores for various image pairs and determined that a five pixel threshold rejects most of these image pairs without also rejecting many valid but challenging pairs. For more limited sequestered testing, this would be done entirely manually to ensure accuracy, and specific SGM failure cases could be added for further assessment. Table 4 shows the results of training and testing for a dozen epochs with this dataset which has far less label noise. Here we see a notable reduction in the fraction of outliers when including semantic label prior information in the stereo correspondence network, as expected. This clearly shows the need for reliable automated image pair selection which is the primary challenge in the multi-view semantic 3D reconstruction challenge track which includes all of these images. For the pairwise semantic stereo track, these image pairs are excluded.

Table 4: Pairwise stereo and semantic stereo performance with image pairs after removing label noise

| Method | EPE | D1 |
| --- | --- | --- |
| SGM + WLS | 3.10 | 0.22 |
| DenseMapNet | 2.15 | 0.20 |
| DenseMapNet + Truth CLS | 1.69 | 0.14 |

## 7. Conclusion

We have presented a novel large-scale public dataset for semantic stereo with multi-view, multi-band, incidental satellite images and an automated methodology for producing labeled challenge data that can be exploited to further extend the dataset. Our initial experiments with lightweight baseline algorithms demonstrate the utility of combining complementary tasks of semantic segmentation and stereo correspondence for improving performance for image pairs with significant seasonal appearance change. The occurrence of geometric change among many incidental image pairs emphasizes the need for reliable image pair selection in multi-view 3D reconstruction. We plan to leverage this dataset to enable a public academic research contest in coordination with the IEEE Geoscience and Remote Sensing (GRSS) Image Analysis and Data Fusion (IADF) technical committee.

Additional work is ongoing to further refine semantic labels and to address remaining issues observed in the data prior to public release. Development is also in work to demonstrate baseline performance for multi-view 3D reconstruction, single-view height estimation, and point cloud semantic segmentation. Baseline implementations will be made publicly available with the data.

## Acknowledgements

The authors are grateful to the IEEE GRSS IADF committee chairs – Bertrand Le Saux, Ronny Hänsch, and Naoto Yokoya – for their collaboration in leveraging this work to enable public research and for important recommendations to improve the challenge tracks. The authors also acknowledge correspondence with and helpful suggestions from Konrad Schindler, Matt Leotta, Joe Mundy, Noah Snavely, and Raquel Urtasun. This work was made possible by the advocacy and support from HakJae Kim.

Commercial satellite imagery was provided courtesy of DigitalGlobe. U. S. Cities lidar and vector data were made publicly available by the Homeland Security Infrastructure Program. Geomni lidar and oblique imagery will be made available publicly for single use research purposes.

This work was supported by the Intelligence Advanced Research Projects Activity (IARPA) contract no. 2017-17032700004. The U.S. Government is authorized to reproduce and distribute reprints for Governmental purposes notwithstanding any copyright annotation thereon. Disclaimer: The views and conclusions contained herein are those of the authors and should not be interpreted as necessarily representing the official policies or endorsements, either expressed or implied, of IARPA or the U.S. Government.